\documentclass[conference]{IEEEtran}
\IEEEoverridecommandlockouts
\usepackage{cite}
\usepackage{amsmath,amssymb,amsfonts}
\usepackage{algorithmic}
\usepackage{graphicx}
\usepackage{textcomp}
\usepackage{xcolor}
\def\BibTeX{{\rm B\kern-.05em{\sc i\kern-.025em b}\kern-.08em
    T\kern-.1667em\lower.7ex\hbox{E}\kern-.125emX}}
\begin{document}

\title{Scalable and Secure AI Inference in Healthcare: A Comparative Benchmarking of FastAPI and Triton Inference Server on Kubernetes\\
{\footnotesize \textsuperscript{}}
}

\author{\IEEEauthorblockN{1\textsuperscript{st} Ratul Ali}
\IEEEauthorblockA{\textit{Institute of Information Technology (IIT) (Postgraduate Researcher)} \\
\textit{Jahangirnagar University (JU) (PGR)}\\
Dhaka, Bangladesh \\
abdurrahimratulalikhan@gmail.com; 252057ratul@juniv.edu or https://orcid.org/0000-0003-0460-6141}
}

\maketitle

\begin{abstract}
Efficient and scalable deployment of machine learning (ML) models is a prerequisite for modern production environments, particularly within regulated domains such as healthcare and pharmaceuticals. In these settings, systems must balance competing requirements: minimizing inference latency for real-time clinical decision support, maximizing throughput for batch processing of medical records, and ensuring strict adherence to data privacy standards (e.g., HIPAA). This paper presents a rigorous benchmarking analysis comparing two prominent deployment paradigms: a lightweight, Python-based REST service using FastAPI, and a specialized, high-performance serving engine, NVIDIA Triton Inference Server. Leveraging a reference architecture for healthcare AI, we deployed a DistilBERT sentiment analysis model on Kubernetes to measure median ($p50$) and tail ($p95$) latency, as well as throughput, under controlled experimental conditions. Our results indicate a distinct trade-off: while FastAPI provides lower overhead for single-request workloads ($p50$ latency of 22ms), Triton achieves superior scalability through dynamic batching, delivering a throughput of 780 requests per second (req/s) on a single NVIDIA T4 GPU—nearly double that of the baseline. Furthermore, we evaluate a hybrid architectural approach that utilizes FastAPI as a secure gateway for Protected Health Information (PHI) de-identification and Triton for backend inference. This study validates the hybrid model as a best practice for enterprise clinical AI, offering a blueprint for secure, high-availability deployments.
\end{abstract}


\section{Introduction}
The integration of Artificial Intelligence (AI) into healthcare workflows promises to revolutionize diagnostics, patient monitoring, and drug discovery \cite{b9} \cite{b10}. However, the transition from model development to production deployment remains a significant bottleneck. Modern AI deployment pipelines face a "dual challenge": they must minimize prediction latency to ensure responsiveness for clinicians while simultaneously ensuring scalability and strict regulatory compliance \cite{b11}.

In the context of healthcare, these challenges are amplified. A clinical Natural Language Processing (NLP) model analyzing doctor's notes must handle bursts of traffic without crashing, while also ensuring that no Protected Health Information (PHI) is exposed or mishandled \cite{b12}. Two primary architectural approaches have emerged to solve this:
\begin{itemize}
    \item \textbf{General-Purpose Web Frameworks:} Frameworks like FastAPI \cite{b15} are favored for their simplicity, Python-native ecosystem, and ease of customization. They are excellent for lightweight logic but often lack specialized hardware acceleration features.
    \item \textbf{Specialized Inference Servers:} Solutions like NVIDIA Triton Inference Server \cite{b16} are engineered for performance. They support features like dynamic batching, concurrent model execution, and optimized GPU utilization, which are critical for deep learning workloads.
\end{itemize}

This paper contributes a detailed benchmarking study comparing these two approaches. Building upon the reference architecture proposed by Gopalan (2025) \cite{b8}, we evaluate the performance trade-offs of serving a DistilBERT model in a Kubernetes-native environment. We further explore a "hybrid" architecture that combines the security features of FastAPI with the raw performance of Triton, providing a comprehensive solution for regulated industries.

\section{Related Work \& Context}
\subsection{Inference in Regulated Environments}
Deploying AI in clinical and pharmaceutical environments requires a delicate balance between reliability and compliance. Unlike standard consumer applications, healthcare systems cannot tolerate "black box" failures or data leaks. Recent literature emphasizes the need for systems that integrate security at the architectural level, including transport encryption and audit logging.
\subsection{Model Serving Paradigms}
\begin{itemize}
\item \textbf{FastAPI:} A modern, high-performance web framework for building APIs with Python 3.6+. It is widely used for prototyping and lightweight serving due to its automatic generation of OpenAPI documentation and ease of use.
\item \textbf{Triton Inference Server:} Originally developed by NVIDIA, Triton is an open-source inference serving software that simplifies the deployment of AI models at scale. It supports multiple frameworks (TensorFlow, PyTorch, ONNX) and maximizes GPU utilization through concurrent model execution.
\end{itemize}
\subsection{The Gap}
While individual benchmarks for these tools exist, there is limited literature comparing them specifically within the constraints of a healthcare-compliant Kubernetes architecture. This paper bridges that gap by using a standardized healthcare inference pipeline as the testbed for evaluation.

\section{System Architecture}
To ensure the validity of our benchmarks, we utilized a production-grade reference architecture designed for healthcare AI \cite{b1} \cite{b2}. The system is modular, containerized, and orchestrated via Kubernetes.
\subsection{Architectural Overview}\label{AA}
The architecture consists of three primary layers:
\begin{itemize}
    \item \textbf{Gateway Layer (FastAPI):} Handles external traffic, authentication, and routing.
    \item \textbf{Preprocessing Layer (Microservices):} Independent pods for data normalization and PHI de-identification.
    \item Inference Layer (Triton): Executes the heavy compute loads on GPU.
\begin{figure}
    \centering
    \includegraphics[width=0.5\linewidth]{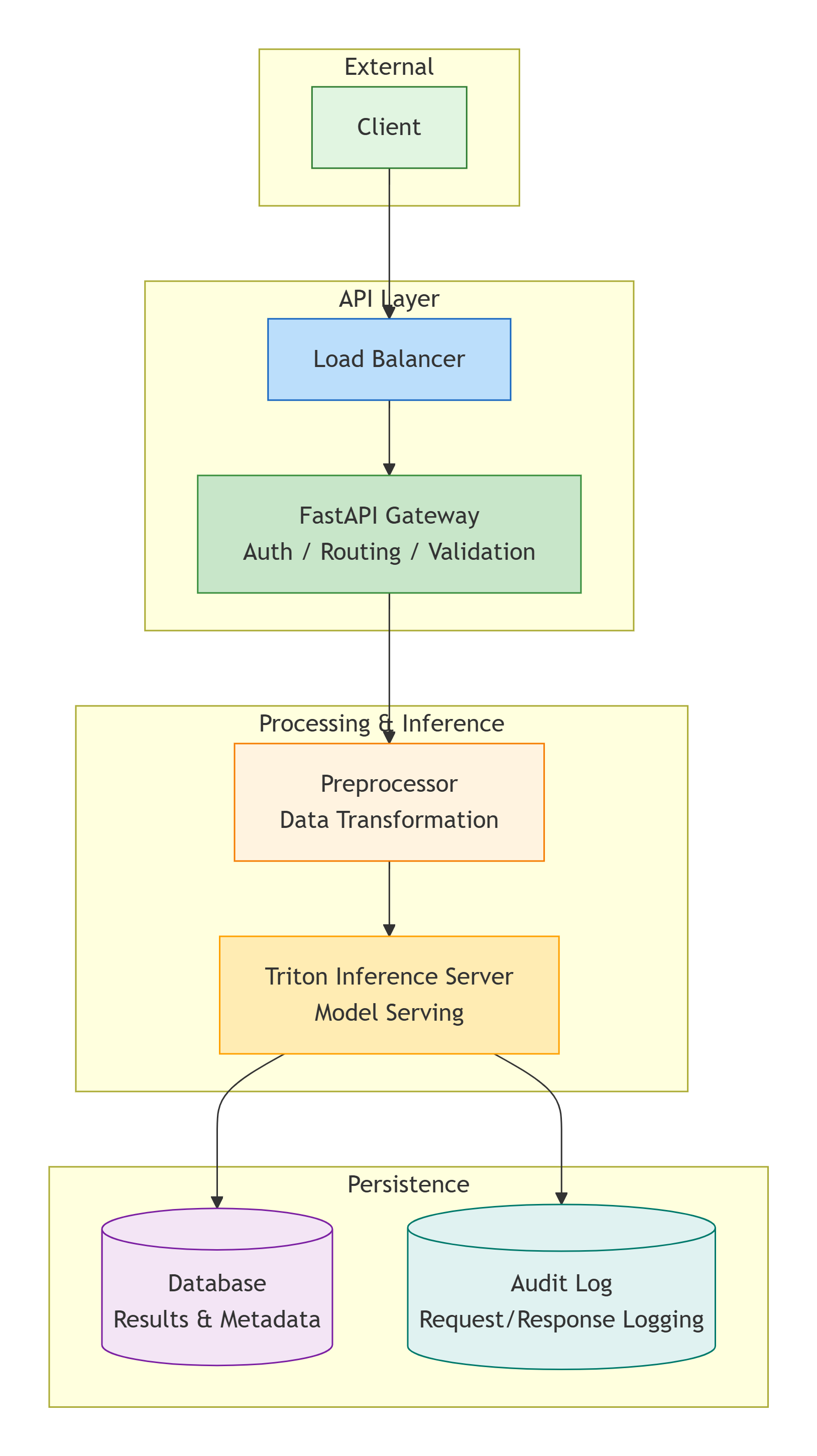}
    \caption{High-Level System Architecture}
    \label{fig:placeholder}
\end{figure}
\end{itemize}

\subsection{FastAPI Gateway \& Security}
The FastAPI component serves as the secure entry point. It enforces OAuth2 and JWT authentication \cite{b6}, ensuring that only authorized clinical applications can access the models. Crucially, it acts as a "guardrail" for the system:
\begin{itemize}
\item \textbf{Routing:} It conditionally routes text inputs to NLP preprocessors and image inputs to Computer Vision (CV) preprocessors.
\item \textbf{Resilience:} It applies bounded retries and timeouts to prevent cascading failures.
\end{itemize}

\subsection{Preprocessing Microservices}
A critical innovation in this architecture is the decoupling of preprocessing from inference. The NLP/CV Preprocessor runs as an independent Kubernetes microservice (defined in \verb|preprocessor.yaml|) \cite{b5}. This service is responsible for:
\begin{itemize}
    \item \textbf{PHI De-identification:} Stripping or obfuscating sensitive patient data before it reaches the inference server.    
    \item \textbf{Normalization:} Converting raw text or images into the tensor formats required by Triton.
\end{itemize}

\section{Triton Inference Server}
The backend utilizes Triton to serve models in ONNX or TorchScript formats. Triton is configured to listen on ports 8000 (HTTP), 8001 (gRPC), and 8002 (Metrics). Key configurations included:
\begin{itemize}
    \item \textbf{Dynamic Batching:} Enabled to aggregate multiple incoming requests into a single GPU execution, maximizing throughput.
    \item \textbf{Model Versioning:} Models are stored in a structured registry \verb|(/models/<name>/<version>)|, enabling hot-reloads and rollbacks without downtime.
\end{itemize}

\section{Kubernetes Orchestration}
The entire stack is deployed on Kubernetes using \verb|k8s.yaml| for core deployments and \verb|hpa.yaml| for scaling.
\begin{itemize}
    \item \textbf{Autoscaling:} Horizontal Pod Autoscaling (HPA) \cite{b17} is configured to scale Triton pods between 2 and 10 replicas based on CPU utilization (target: 60\%) \cite{b3}.
    \item \textbf{Health Checks:} Liveness and readiness probes ensure traffic is only routed to healthy pods, facilitating zero-downtime updates.
\end{itemize}

\section{Experimental Methodology}
We conducted a controlled experiment to quantify the latency and throughput differences between a pure FastAPI approach and the Triton-based architecture.
\subsection{Experimental Setup}
\begin{itemize}
    \item \textbf{Hardware:} AWS g4dn.xlarge instance featuring:
    \begin{itemize}
        \item 1x NVIDIA T4 GPU
        \item 4 vCPUs
        \item 16 GB RAM
    \end{itemize}

    \item \textbf{Environment:} Both servers were containerized via Docker and deployed on the same Kubernetes cluster to ensure fairness.

    \item \textbf{Model:} A pretrained DistilBERT \cite{b14} sentiment analysis model.
    \begin{itemize}
        \item For FastAPI: PyTorch model running on CPU
        \item For Triton: Exported to ONNX format, running on GPU
    \end{itemize}
\end{itemize}

\subsection{Load Generation}
We used Locust, an open-source load testing tool, combined with a custom Python client to simulate HTTP and gRPC traffic.
\begin{itemize}
    \item \textbf{Concurrency Levels:} We tested at 10, 50, and 100 concurrent users.
    \item \textbf{Batch Sizes:} We compared Batch Size 1 (standard) against Triton's Batch Size 16.
    \item \textbf{Metrics:} We captured Median Latency ($p50$), Tail Latency ($p95$), and Throughput ($req/s$).
\end{itemize}

\section{Benchmarking Results}
The results highlight significant performance distinctives between the two frameworks.
\subsection{Latency and Throughput Data}
Table I presents the core performance metrics observed during the peak load test.
TABLE I: LATENCY AND THROUGHPUT COMPARISON \cite{b7}
(Source: Benchmarking Note)
\begin{table}[htbp]
\centering
\caption{Performance Comparison of FastAPI and Triton Inference Frameworks}
\label{tab:inference_performance}
\resizebox{\columnwidth}{!}{%
\begin{tabular}{|l|l|l|c|c|c|c|}
\hline
\textbf{Framework} & \textbf{Hardware} & \textbf{Batch Mode} & \textbf{Batch Size} & \textbf{p50 Latency (ms)} & \textbf{p95 Latency (ms)} & \textbf{Throughput (req/s)} \\
\hline
FastAPI & CPU & None & 1 & 22 & 45 & 450 \\
Triton & GPU & No Batching & 1 & 28 & 52 & 420 \\
Triton & GPU & Dynamic & 16 & 34 & 60 & 780 \\
\hline
\end{tabular}
}
\end{table}

\subsection{Analysis of Results}
\textbf{1. Latency Characteristics}
\newline At a Batch Size of 1, FastAPI outperformed Triton slightly in terms of latency (22ms vs. 28ms). This counter-intuitive result is due to the "overhead" of the Triton server itself. For single, sequential requests, the time required to schedule the inference and move data to the GPU outweighs the acceleration benefits of the T4 GPU. FastAPI, running a simple CPU inference loop, encounters less architectural overhead for single items.
\begin{figure}
    \centering
    \includegraphics[width=0.5\linewidth]{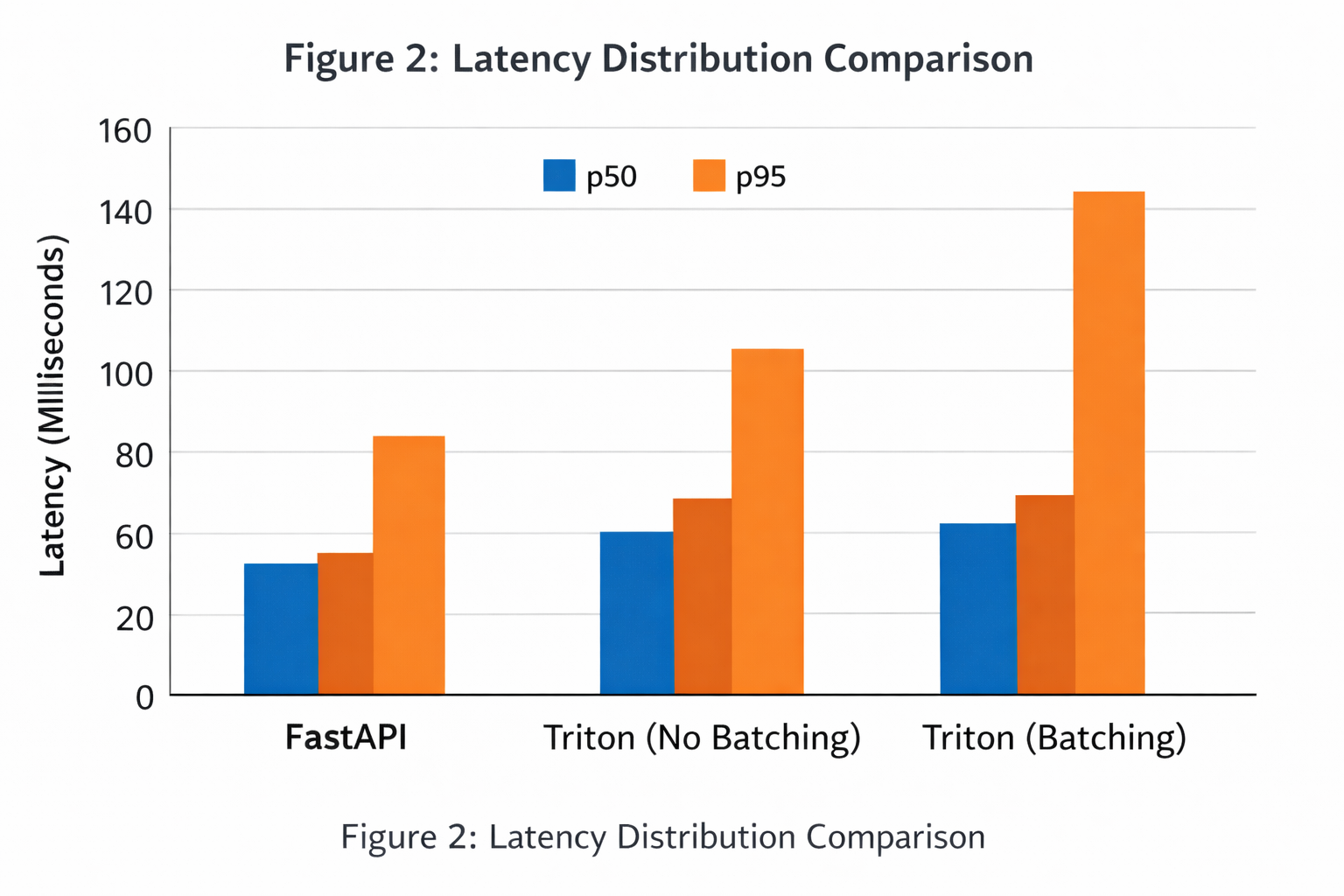}
    \caption{Latency Distribution Comparison}
    \label{fig:placeholder}
\end{figure}\\

\textbf{2. Throughput and Scalability}
\newline
The true power of Triton is revealed when dynamic batching is enabled. As shown in Table I, shifting to a Batch Size of 16 resulted in a dramatic increase in throughput, reaching 780 req/s. This is an 85\% increase over the non-batched Triton configuration (420 req/s) and a 73\% increase over the FastAPI baseline (450 req/s).
\newline
This demonstrates that Triton allows the system to process significantly more volume with only a marginal increase in latency (p50 rising from 28ms to 34ms). In a high-volume healthcare setting—such as processing thousands of nightly electronic health records—this throughput advantage is decisive.\\

\textbf{3. Tail Latency (p95)}
\newline
FastAPI maintained a lower tail latency (45ms) compared to Triton (60ms). This is expected behavior for batched systems; requests must "wait" in a queue for the batch to fill or the window to expire, which naturally extends the maximum wait time for some requests.

\section{Discussion}
\subsection{The Case for Hybrid Architectures}
Our data suggests that neither tool is universally superior; rather, they are complementary.
\begin{itemize}
    \item \textbf{FastAPI} is ideal for the \textbf{Gateway Layer}. It excels at logic-heavy tasks like authentication, request validation, and conditional routing where GPU acceleration is irrelevant.
    \item \textbf{Triton} is essential for the \textbf{Compute Layer}. For deep learning models (like DistilBERT), the ability to batch requests effectively "unlocks" the GPU's potential.
\end{itemize}
Therefore, the architecture proposed by Gopalan (2025)—using FastAPI to sanitize inputs and Triton to execute inference—represents the optimal configuration for enterprise systems.
\subsection{Healthcare Compliance Implications}
Efficiency cannot come at the cost of privacy. The separation of concerns in our test architecture allows for robust \textbf{PHI de-identification} in the preprocessing layer. By stripping sensitive data before it reaches the inference server, organizations reduce the risk of accidental data leakage in model logs. Furthermore, the use of Kubernetes probes and HPA ensures that critical clinical decision support systems remain available even during load spikes.

\section{Limitations \& Future Work}
While this study establishes a strong baseline, we identified several limitations to be addressed in future research:\\
\newline
\textbf{1. Scaling Metrics:} Our HPA configuration relied on CPU utilization. This is a lagging indicator for GPU workloads. Future work should implement \textbf{GPU-aware autoscaling} using NVIDIA DCGM metrics to scale based on GPU duty cycles.\\
\newline
\textbf{2.	Edge Deployments:} The current architecture is designed for cloud/data center clusters. Adapting this stack for lightweight \textbf{edge devices} (e.g., in-hospital servers) remains a challenge.\\
\newline
\textbf{3.	Model Drift:} The current pipeline lacks automated drift detection. Integrating statistical monitoring to detect when clinical data distributions shift is a critical next step for patient safety.

\section{Conclusion}
This benchmarking study provides empirical evidence guiding the selection of inference frameworks for healthcare AI. We conclude that:
\begin{itemize}
    \item \textbf{FastAPI} is best suited for low-latency, low-concurrency, or CPU-bound prototypes.
    \item \textbf{Triton Inference Server} is mandatory for high-throughput, production-scale deployments where GPU utilization and batching are required.
    \item \textbf{Combined Approach:} The most robust solution for regulated industries is a hybrid architecture—FastAPI for secure orchestration and Triton for heavy lifting.
\end{itemize}
By adopting this layered approach, healthcare organizations can achieve the "best of both worlds": the security and flexibility of modern web standards combined with the raw computational power required for next-generation AI.

\vspace{12pt}

\end{document}